\documentclass{llncs}
\usepackage[T1]{fontenc}
\usepackage{graphicx}
\usepackage{scrextend}
\usepackage{bigfoot}
\usepackage[splitrule,bottom]{footmisc} 
\usepackage{multirow}
\usepackage{times}
\usepackage{latexsym}

\usepackage[acronym]{glossaries}
\newacronym{bp}{BP}{Brazilian Portuguese}
\newacronym{ep}{EP}{European Portuguese}
\newacronym{nlp}{NLP}{Natural Language Processing}
\newacronym{lstm}{LSTM}{Long Short-Term Memory}
\newacronym{nmt}{NMT}{Neural Machine Translation}
\newacronym{rnn}{RNN}{Recurrent Neural Network}
\newacronym{mt}{MT}{Machine Translation}
\newacronym{gpt}{GPT}{Generative Pre-trained Transformer}
\newacronym{k}{K}{Thousand}
\newacronym{mqm}{MQM}{Multidimensional Quality Metrics}
\newacronym{llm}{LLM}{Large Language Model}
\newacronym{api}{API}{Application Programming Interface}

\usepackage{color}

\begin{document}
\title{From Brazilian Portuguese to European Portuguese}
\titlerunning{BR2PT}
%
\author{Jo\~{a}o Sanches \and
Rui Ribeiro \and
Luisa Coheur}

\authorrunning{J. Sanches et al.}
%
\institute{Instituto Superior T\'{e}cnico, Universidade de Lisboa/INESC-ID\\ 
\email{\{joao.sanches,rui.m.ribeiro,luisa.coheur\}@tecnico.ulisboa.pt}}
\maketitle              
\begin{abstract}
Brazilian Portuguese and European Portuguese are two varieties of the same language and, despite their close similarities, they exhibit several differences. However, there is a significant disproportion in the availability of resources between the two variants, with Brazilian Portuguese having more abundant resources. This inequity can impact the quality of translation services accessible to European Portuguese speakers. To address this issue, we propose the development of a Brazilian Portuguese to European Portuguese translation system, leveraging recent advancements in neural architectures and models.
To evaluate the performance of such systems, we manually curated a gold test set comprising 500 sentences across five different topics. Each sentence in the gold test set has two distinct references, facilitating a straightforward evaluation of future translation models. We experimented with various models by fine-tuning existing Large Language Models using parallel data extracted from movie subtitles and TED Talks transcripts in both Brazilian and European Portuguese. Our evaluation involved the use of conventional automatic metrics as well as a human evaluation. In addition, all models were compared against ChatGPT 3.5 Turbo, which currently yields the best results.  
\end{abstract}


\section{Introduction}
Despite \gls*{bp} and \gls*{ep} being very similar, linguistic differences between \gls*{ep}-speakers and \gls*{bp}-speakers, such as orthography and syntax \cite{mateusalii}, may detriment their mutual understanding. However, services that offer translation between \gls*{bp} and \gls*{ep} have limited capabilities. For instance, until recently Google Translate\footnote{\url{https://translate.google.com}} only offered translation to a standardized Portuguese\footnote{\url{https://cloud.google.com/translate/docs/languages}}, while DeepL \cite{yulianto2021google}, despite offering translation for \gls*{bp} and \gls*{ep} to other languages, does not allow to directly translate between  \gls*{bp} and \gls*{ep}.

In this paper, we tackle the challenge of translating from \gls*{bp} to \gls*{ep}. We will focus on \gls*{nmt}-based models and delve into the process of fine-tuning multilingual models, namely M2M100 418M\footnote{\url{https://huggingface.co/facebook/m2m100\_418M}} and mBART-large-50\footnote{\url{https://huggingface.co/facebook/mbart-large-50}}; we also tested ChatGPT 3.5 Turbo\footnote{\url{https://chat.openai.com}}.

Furthermore, we also make available our training, development, and test datasets, which were collected from TED Talks\footnote{\url{https://www.ted.com/}} transcripts and subtitles from OPUS \cite{lisontiedemann2016opensubtitles2016}. 
We also contribute with a manually curated gold dataset of 500 parallel \gls*{bp}-\gls*{ep} sentences, from 5 different domains. This collection, which we also make available, allows for direct comparison between these and future models performing this task.

Our results show that the mBART-large-50 model, fine-tuned with a mixed dataset of TED Talks and subtitles, performs best among our models. However, it falls slightly short of surpassing ChatGPT 3.5 in the Golden Collection, despite achieving comparable performance.

This paper is organized as follows: in Section~\ref{sec:rw} we present related work, in Section~\ref{sec:datasets} we introduce the manual curated Golden Collection, and in Section~\ref{sec:training} we detail the experimental setup. In Section~\ref{sec:Evaluation} we present and discuss the obtained results, and, finally, in Section~\ref{sec:conc} we highlight the main conclusions and point to future work.


\section{Related Work} \label{sec:rw}

The first translation tasks were performed by human translators, relying on their linguistic expertise, cultural understanding, and contextual knowledge to ensure accurate and culturally relevant translations \cite{altintas2022machine}. Then, \gls*{mt} emerged, and different translation architectures appeared: first rule-based translation, then statistical \gls*{mt}, and lastly \gls*{nmt} \cite{poibeau2017machine}, which currently represents the state-of-art in this area. 

To the best of our knowledge, the first work that proposes a translation between \gls*{bp} and \gls*{ep}, is presented in \cite{marujo2011bp2ep}. This work takes advantage of TED Talks to build phrase tables in a classic \gls*{mt} framework, however, it is no longer available.  

A more recent work, is the one described in \cite{costa2018neural}. In this work, the authors employ a \gls*{rnn} encoder-decoder architecture with an attention mechanism \cite{costa-jussa-etal-2017-byte}. More specifically, an improvement of \glspl*{rnn} with enhanced memorization abilities, a \gls*{lstm}, is responsible for the encoding phase, while a \gls*{rnn} is responsible for decoding the target translation text. 

Recently, the work described in \cite{adebara2021improving} developed various systems for translating between Catalan-Spanish, Portuguese-Spanish, and Franch-Bambara\footnote{\url{https://huggingface.co/Ife}} language pairs. This work uses various models based on Transformers \cite{TRANSFORMERS}, which is the base architecture used in current state-of-the-art pre-trained models, such as BERT \cite{BERT} and \gls*{gpt} \cite{GPT}. The authors leveraged these large models by fine-tuning with the language pairs from the target dataset for the \gls*{nmt} task. By using this transfer learning method, it is possible to help mitigate some data scarcity issues that happen in the real-world scenario. Additionally, this work found that using fine-tuned models could enhance the performance relative to models trained from scratch and that the models with tokenized data outperformed the ones without tokenization. 

Other works explore the multilingual setting. For instance, in \cite{johnson2017google}, the authors introduce a different approach by employing a multilingual \gls*{nmt} model where all parameters are shared across multiple language pairs, enabling the model to generalize across language boundaries during training. 

Our work is inspired by the study conducted by \cite{cortes-etal-2024-llms}, which investigates the effectiveness of \gls*{mt} approaches for adapting linguistic and cultural material between \gls*{ep} and \gls*{bp}. 
The study assesses the extent to which MT models paraphrase sentences during the process of translation and evaluates their performance using both human and automatic evaluations. Four models were tested and analyzed, including a rule-based with a masked language model, a fine-tuned pre-trained \gls*{nmt}, and two \gls*{gpt}-4-based models. We adopt a similar strategy with some modifications.
Our approach focuses on training and evaluating a wider variety of fine-tuned pre-trained models. The pre-trained model used in their study was mBART-50 with 100,000 sentences from OpenSubtitles \cite{lisontiedemann2016opensubtitles2016}, similar to one of our experiments. 

Lastly, evaluating the performance of applications based on Large Language Models, such as ChatGPT, is also essential.  In \cite{jiao2023chatgpt}, the authors introduce a study that concludes that ChatGPT (built upon GPT-3) performs competitively with commercial translation products (e.g., Google Translate), particularly in well-resourced European languages. In addition, in \cite{hendy2023good}, the authors demonstrate that \gls*{gpt}-3 based systems excel in high-resource language translations, being able to achieve state-of-art translation when combined with \gls*{nmt} systems. However, the authors also found that ChatGPT is still not as effective with respect to low-resource distant languages. In \cite{jiao2023chatgpt}, it is suggested that ChatGPT does not perform as well in domain-specific areas and noisy environments, such as Reddit.

\section{Towards the Gold Collection} \label{sec:datasets}


\subsection{Gathering the Data}

The Golden Collection comprises 500 aligned sentence pairs, bridging \gls*{bp} and \gls*{ep}. These pairs are categorized into five distinct domains: scientific, legislative, social media, literature, and wiki (100 pairs each). Sentences pertaining to the social media, legislative, and wiki domains were extracted from the Carolina corpus\footnote{\url{https://huggingface.co/datasets/carolina-c4ai/corpus-carolina}}, which contains contemporary \gls*{bp} sentences of varied typology \cite{crespo2023carolina}.  
Sentences from the scientific domain were gathered from Summ-it\footnote{\url{https://portulanclarin.net/repository/extradocs/Summit.pdf}}, a repository containing 50 scientific texts extracted from the Brazilian newspaper Folha de São Paulo; and finally we extracted literary sentences from \gls*{bp} books from the website Baixe Livros\footnote{\url{https://www.baixelivros.com.br/biblioteca/literatura-brasileira}}.

The selection of sentences from each corpus was performed randomly. However, due to the high level of noise in some datasets, such as those from social media, we removed emojis and other noisy symbols. For instance, in the wiki domain, which had metadata in the main corpus, we removed the sentence ``[...] LaViHD@C4AI 2021-08-04 2022-09-22 CC BY-SA 4[ ...]". Moreover, we removed sentences in other languages. 

Table \ref{tab:Gold} shows statistics of the Golden Collection. We can observe that the corpus is composed of a wide range of sentence lengths and unique word tokens across and within each category. The category that has the lowest value in size and diversification is the social media category, while at the other end of the spectrum, we have the scientific category. This is expected because sentences from social media sources are meant to be quickly read, therefore they will be smaller. On the other hand, scientific texts are usually more verbose. 

\begin{table*}
    \centering
    \caption{Golden Dataset statistics with respect to the number of word tokens.} 
    \begin{tabular}{c|c|c|c|c|c|c} \hline
         &  Total & Unique &  Median & Average & Min & Max\\ \hline 
         Scientific&  2,516 & 1,137 &  24 & 24.9 & 4 & 60\\ 
         Legislative&  2,362 & 911 & 20 & 23.6 & 3 & 126\\ 
         Literature&  1,704 & 765 &  12 & 16.9 &  3 & 105\\ 
         Social Media&  1,434 & 675 & 12  & 14.1 & 3 & 63\\ 
         Wiki&  2800 & 1,290 & 25  &  28.0 & 6 & 76\\ 
         Golden (Total)& 10,807 & 3,722 & 18.0 &  21.6 & 3 & 126\\  \hline
     \end{tabular}
    \label{tab:Gold}
\end{table*}

\begin{table*}[!ht]
\centering
\caption{Golden Collection examples.} 
\begin{tabular}{ l | p{0.75\textwidth}}
\hline
Category  & Examples \\
\hline
\multirow{3}{*}{Science}  & \textbf{BP:} Desse total, 200 milhões se devem ao desmatamento.\\ 
& \textbf{EP - Manual:} Desse total, 200 milhões devem-se à desflorestação.  \\ 
& \textbf{EP - DeepL:} Deste total, 200 milhões devem-se à desflorestação.\\
& \textbf{EN:}  From this total, 200 million are due to deforestation. \\
\hline
\multirow{3}{*}{Legislative} & \textbf{BP:} Parágrafo único – Considera se atividade econômica atípica aquela realizada no recesso do lar.\\ 
& \textbf{EP - Manual:} Parágrafo único -- Considera-se atividade económica atípica aquela realizada no refúgio do lar.\\ 
& \textbf{EP - DeepL:} Parágrafo único -- As actividades económicas atípicas são as exercidas na privacidade do domicílio.\\
& \textbf{EN:} Unique paragraph: It is considered atypical economic activity the one which is performed at home.\\
\hline
\multirow{3}{*}{Social Media} & \textbf{BP:} E so para me respeita garotinha.\\ 
& \textbf{EP - Manual:} É só para me respeitares, pequena menina.\\ 
& \textbf{EP - DeepL:} Respeita-me, menina. \\
& \textbf{EN:} It is just to respect me, little girl.\\
\hline
\multirow{3}{*}{Literature} & \textbf{BP:} Lancei-me fora do ônibus; caminhei à direita e à esquerda; andei como um louco até nove horas da noite.\\ 
& \textbf{EP - Manual:} Pus-me fora do autocarro; caminhei à direita e à esquerda; andei como um louco até às nove horas da noite.\\ 
& \textbf{EP - DeepL:} Atirei-me para fora do autocarro; andei para a direita e para a esquerda; andei como um louco até às nove da noite.\\
& \textbf{EN:} I threw myself out of the bus; I walked right and left; I walked like a madman until nine o'clock at night.\\
\hline
\multirow{3}{*}{Wiki} & \textbf{BP:} Foi então que a vestimenta mais feminina que se conhece começou a ganhar forma: o espartilho.\\ 
& \textbf{EP - Manual:} Foi então que a peça de vestuário mais feminina que se conhece começou a ganhar forma: o espartilho.\\ 
& \textbf{EP - DeepL:} Foi nessa altura que começou a ganhar forma a peça de vestuário mais feminina que se conhece: o espartilho.\\
& \textbf{EN:}  It was then that the most feminine garment known to man began to take shape: the corset. \\
\hline
\end{tabular}
\label{tab:examples}
\end{table*}

\subsection{Building the Golden Collection}

After creating the monolingual \gls*{bp} Golden Collection, we built two references in \gls*{ep}. The first reference was created manually, while the second utilized DeepL. For the manually created reference, we made the minimal number of changes necessary to the original text. Since DeepL does not support direct translation between \gls*{bp} and \gls*{ep}, we used English as the pivot language. This resulted in a translation that is significantly more distant from the original text compared to the manually constructed reference. In addition, some manual corrections were applied to the DeepL translation. Table \ref{tab:examples} has examples of sentences of the Golden Collection from each category, alongside the corresponding references. 

\subsection{Evaluating the Golden Collection}

A human evaluator assessed the Golden Collection using the Likert scale \cite{joshi2015likert}. The evaluation involved 50 randomly selected sentences, which were analyzed for both adequacy and fluency. The results are summarized in Table \ref{tab:humanGold}. Notably, the DeepL reference achieved a higher fluency score compared to the manual reference. This may be due to the translation method employed: DeepL made more extensive corrections to the source sentence to enhance fluidity, whereas the manual reference implemented only the minimal changes necessary to achieve the target translation. Nonetheless, the evaluation indicated that both references received high scores for adequacy and fluency, suggesting that both are of high quality and appropriate for evaluating our models.

\begin{table*}[!ht]
\centering
\caption{Human evaluation results.}
\begin{tabular}{l|cc|cc} \hline
 &  \multicolumn{2}{c}{Manual} & \multicolumn{2}{c}{DeepL} \\
\cline{2-5}
\multirow{1}{*}{Domain} & Adequacy & Fluency & Adequacy & Fluency\\
\hline
Scientific &  5.0 & 4.9 & 4.8 & 4.8\\
Legislative &  4.8 & 4.7 & 5.0 & 4.8 \\
Literature &  4.7& 4.3 & 4.3 & 5.0\\
Social Media & 4.8 & 4.2 & 4.7 & 4.8\\
Wiki & 4.9 & 4.7 & 5.0 & 5.0\\
\hline
Average & 4.8 & 4.6 & 4.8 & 4.9 \\ \hline
\end{tabular}
\label{tab:humanGold}
\end{table*}


\section{Experimental Setup}\label{sec:training}

\subsection{Datasets}


No pre-processing was done to the subtitles and TED Talks datasets, as they already consisted of aligned pairs, \gls*{bp}-\gls*{ep}. The resulting datasets (see also Table \ref{tab:statsDatasetsEP}) are:
\begin{itemize}
    \item Subtitles Training Dataset: 126,984 aligned BP-EP subtitle sentences.
    \item TED Talks Training Dataset: 126,984 aligned BP-EP TED Talks sentences.
    \item Merged Training Dataset: Balanced mix of aligned subtitles and TED Talks sentences, with a total of 126,984 sentences, half from each source.
    \item Merged Validation Dataset: Balanced mix of aligned subtitles and TED Talks sentences, with a total of 12,698 sentences. No overlap with training datasets.
    \item Merged Test Dataset: Comprises a balanced mix of aligned subtitles and TED Talks sentences. No overlap with training, evaluation, or test sets is ensured.
\end{itemize}

\begin{table}[ht]
    \centering
    \caption{Datasets statistics with respect to the number of word tokens.}   
    \begin{tabular}{c|c|c|c|c|c|c} 
         \hline 
         &  Total & Unique &  Median & Average& Min & Max\\ 
         \hline 
         Subtitles train dataset&  993,570 & 50,730 &  7 & 7.8 & 1 & 397\\ \hline 
         Ted Talks train dataset&  2 439,714 & 80,106 & 16 & 19.2 & 2& 692\\ \hline 
         Merged train dataset & 1 711,744  & 74,839 & 10 & 13.5 & 1 & 692\\ \hline 
         Merged valid dataset & 167,238 & 17,892 & 9 & 13.2 & 1 & 354\\ \hline 
         Merged test dataset & 163,538 & 17,805 & 9 & 12.9 & 1 & 482\\ 
         \hline
     \end{tabular}
      
    \label{tab:statsDatasetsEP}
\end{table}

\subsection{Models}

Both mBART-50 and M2M100 models employ monolingual pre-training and multilingual fine-tuning. However, mBART-50 uses an English-centric approach, employing English as a pivot language, which can sometimes lead to a bias towards English and potential language context loss. On the other hand, M2M100 is a multilingual sequence-to-sequence model that can translate directly between any pair of 100 languages without relying on English as a pivot language. It avoids potential biases by using language families and bridge languages. M2M100 aims to preserve practical translation directions while reducing the need for numerous bitext pairs. As these multilingual models did not have the distinction between \gls*{bp} and \gls*{ep}, just having a standard ``Portuguese" language, we had to fine-tune these models with parallel corpora in order to make them learn the distinction between these varieties. 

First, we tokenized our training and evaluation data: for that end, the source and target language of the tokenizer for each model was the ``Portuguese" language. After that, both M2M100 and mBART-50 were fine-tuned three times each, once with every training dataset designed for the text-to-text translation.  

Regarding the experimental details, the models were trained for 10 epochs using an NVIDIA A100 80GB GPU. Each training session took approximately 9 to 10 hours. The hyper-parameters used for fine-tuning the pre-trained text-to-text translator models in this paper are as follows: a learning rate of $2\times10^{-5}$, a weight decay of 0.01, training and evaluation batch sizes of 8, and a maximum length of 128 for both the source and target tokenizers. These hyper-parameters were selected to optimize the models' performance during the fine-tuning process.

After this process, we achieved the following models: 
\begin{itemize}
\item Merged M2M100: M2M100 trained with the Merged Training Dataset.
\item TED M2M100: M2M100 trained with the TED Talks dataset.
\item Subtitles M2M100: M2M100 trained with the subtitles dataset.
\item Merged mBART-50: mBART-50 trained with the Merged Training Dataset.
\item TED Talks mBART-50: mBART-50 trained with the TED Talks dataset.
\item Subtitles mBART-50: mBART-50 trained with the subtitles dataset.
\end{itemize}

We also evaluated ChatGPT 3.5 Turbo, which did not receive any training or fine-tuning from us: we used this model in April 2024.

\subsection{Evaluation Metrics}

We performed the evaluation of these models using NLP-Telescope \cite{rita2023telescope}, an accessible interface that can evaluate across a wide range of metrics in any MT model. The metrics chosen were BLEU \cite{papineni2002bleu},  TER \cite{snover-etal-2006-study}, and COMET \cite{rei2020comet}. To have a general score of the performance of the model, NLP-Telescope ranks the models based on a total score, which reflects the overall performance across these metrics, given by:

\begin{equation}
\text{Total Score} = \text{COMET} + \text{BLEU} + (1 - \text{TER}).
\label{eq:score}
\end{equation}

It is also important to note that the COMET score can exceed 1. This happens for two reasons:
\begin{itemize}
\item The version of COMET we are using is COMET-2020-da, which is unbounded. 
\item \gls*{ep} and \gls*{bp} are very similar variants, which leads to very close source, reference, and translation texts. 
\end{itemize}


\section{Results and Discussion}\label{sec:Evaluation}

\subsection{Cross Domain Results}

We kept records of the evaluation loss during the training of each fine-tuned model and selected the best models based on this metric. The M2M100 model fine-tuned with a merged dataset or with just TED Talks achieved the best evaluation loss at epoch 3. Meanwhile, the M2M100 model fine-tuned with subtitles and the mBART-50 model trained with a merged dataset or subtitles reached their lowest evaluation loss at epoch 2. Finally, the mBART-50 model trained with TED Talks minimized the evaluation loss at epoch 1. The best results for the different models evaluated on the various subtitles and TED Talks tests are presented in Table \ref{tab:averages}.

Overall, these scores are not unexpected: fine-tuning a model to a certain domain leads to it being particularly good at translating that domain, which happened repeatedly in this part of the experiment. It is interesting to note that the merged datasets were very close to the score of the top models or even constituted that top score: as they were trained with data from both sources, they became proficient at tasks involving either type of data. As such, the models that seem the most promising so far are the Merged mBART-50 and the Merged M2M100. Regarding the models fine-tuned with TED Talks and the models fine-tuned with subtitles, we conclude that training solely with subtitles leads to the model generalizing better to new data than training only with TED Talks.

\begin{table*}[htb]
\centering
\caption{Test Set: Subtitles, TED Talks, and average results.}
\begin{tabular}{l|rccccc} \hline
Domain & Translation Model & COMET & BLEU & TER & Total Score & Rank\\
\hline
\multirow{7}{*}{TED Talks} 
& MERGED M2M100& 0.80 & 0.38 & 0.50 & 1.68 & 4 \\
& TED M2M100& 0.79 & \textbf{0.40} & \textbf{0.48} & \textbf{1.71} & 1 \\
& SUBS M2M100& 0.79 & 0.32 & 0.56 & 1.55 & 6 \\
& MERGED MBART-50 & 0.80 & 0.38 & 0.49 & 1.69 & 2 \\
& TED MBART-50 & 0.80 & 0.38 & 0.49 & 1.69 & 2 \\
& SUBS MBART-50 & 0.80 & 0.31 & 0.56 & 1.55 & 6 \\
& ChatGPT 3.5 Turbo & \textbf{0.83} & 0.35 & 0.53 & 1.65 & 5 \\
\hline
\multirow{7}{*}{Subtitles} 
& MERGED M2M100 & 0.89 & 0.59 & 0.36 & 2.12 & 4 \\
& TED M2M100 & 0.76 & 0.44 & 0.47 & 1.73 & 7 \\
& SUBS M2M100 & \textbf{0.90} & 0.60 & 0.35 & 2.15 & 2 \\
& MERGED MBART-50 & \textbf{0.90} & 0.60 & 0.35 & 2.15 & 2 \\
& TED MBART-50 & 0.81 & 0.50 & 0.50 & 1.81 & 6 \\
& SUBS MBART-50 & \textbf{0.90} & \textbf{0.61} & \textbf{0.34} & \textbf{2.17} & \textbf{1} \\
& ChatGPT 3.5 Turbo & 0.83 & 0.56 & 0.41 & 1.98 & 5 \\
\hline
\multirow{7}{*}{Average} 
& MERGED M2M100 & \textbf{0.85} & 0.49 & 0.43 & 1.91 & 2 \\
& TED M2M100 & 0.78 & 0.42 & 0.48 & 1.72 & 7 \\
& SUBS M2M100 & 0.84 & 0.46 & 0.45 & 1.85 & 4 \\
& MERGED MBART-50 & \textbf{0.85} & \textbf{0.49} & \textbf{0.42} & \textbf{1.92} & \textbf{1} \\
& TED MBART-50 & 0.80 & 0.44 & 0.50 & 1.74 & 6 \\
& SUBS MBART-50 & \textbf{0.85} & 0.46 & 0.45 & 1.86 & 3 \\
& ChatGPT 3.5 Turbo & 0.83 & 0.46 & 0.47 & 1.82 & 5 \\
\hline
\end{tabular}
\label{tab:averages}
\end{table*}

\subsection{Results in the Gold Collection}

Next, we evaluated the Golden Collection (Table \ref{tab:gold}). An additional experience was conducted, as shown in this table, to evaluate the obtained scores when the translation is exactly the same as the source text (in the table, lines ``Nothing``).  Our main conclusions are the following: 
\begin{itemize}
\item Against the manual reference, our Merged and Subtitles fine-tuned models beat the total score of ChatGPT 3.5 Turbo. Models fine-tuned with TED Talks presented consistently the worst results out of all the models. 
\item Against the DeepL reference, we observed that our fine-tuned models did not beat ChatGPT 3.5 Turbo. ChatGPT was consistently the best model across all metrics. As for the results of our fine-tuned models, there was not much change: models fined-tuned with TED Talks, despite having relatively slightly better scores in comparison to the results against the manual reference, were still part of the bottom scoring models. This indicates that the translation that ChatGPT generates is the most similar to the one generated by DeepL, which is remarkably different from the source text. Our models, on the other hand, fare better with traditional translation, which deviates just a little from the source text.
\item Finally, it is worth mentioning that doing ``nothing" to the source text many times was better than any of our models and ChatGPT. This happens because \gls*{ep} and \gls*{bp} are very similar varieties of the same language. The only instance where the overall score of doing nothing was surpassed was against the DeepL reference. This result is expected, as the DeepL reference is the most divergent from the source text, whereas we aimed to make minimal changes in the manual reference.

\end{itemize}

\begin{table*}[htb]
\centering
\caption{Golden Collection: Manual, DeepL, and Average results.}
\begin{tabular}{r|cccccc} \hline
Translation Model & COMET & BLEU & TER & Total Score & Rank & Reference \\
\hline
MERGED M2M & 1.16 & 0.70 & 0.21 & 2.65 & 6 & Manual \\
TED M2M & 1.03 & 0.62 & 0.27 & 2.38 & 8 & Manual \\
SUBS M2M & 1.17 & 0.84 & 0.10 & 2.91 & 2 & Manual \\
MERGED MBART & 1.17 & 0.84 & 0.11 & 2.90 & 3 & Manual \\
TED MBART & 1.08 & 0.70 & 0.20 & 2.58 & 7 & Manual \\
SUBS MBART & 1.09 & 0.83 & 0.12 & 2.80 & 5 & Manual \\
ChatGPT 3.5 Turbo & 1.16 & 0.84 & 0.12 & 2.88 & 4 & Manual \\
\hline
Nothing & \textbf{1.21} & \textbf{0.89} & \textbf{0.07} & \textbf{3.03} & 1 & Manual \\
\hline
MERGED M2M & 0.93 & 0.50 & 0.37 & 2.06 & 3 & DeepL \\
TED M2M & 0.86 & 0.48 & 0.39 & 1.95 & 8 & DeepL \\
SUBS M2M & 0.95 & 0.49 & 0.38 & 2.06 & 3 & DeepL \\
MERGED MBART & 0.95 & 0.51 & 0.37 & 2.09 & 2 & DeepL \\
TED MBART & 0.92 & 0.51 & 0.37 & 2.06 & 3 & DeepL \\
SUBS MBART & 0.88 & 0.48 & 0.40 & 1.98 & 7 & DeepL \\
ChatGPT 3.5 Turbo & 0.99 & \textbf{0.53} & \textbf{0.36} & 2.16 & 1 & DeepL \\
\hline
Nothing & \textbf{1.00} & 0.49 & 0.38 & 2.11 & 2 & DeepL \\
\hline
MERGED M2M & 1.05 & 0.60 & 0.29 & 2.36 & 6 & Average\\
TED M2M & 0.95 & 0.55 & 0.33 & 2.17 & 8 & Average\\
SUBS M2M & 0.99 & 0.67 & 0.24 & 2.42 & 4 & Average\\
MERGED MBART & 0.99 & 0.68 & 0.24 & 2.43 & 3 & Average\\
TED MBART & 1.00 & 0.60 & 0.29 & 2.31 & 7 & Average\\
SUBS MBART & 0.99 & 0.66 & 0.26 & 2.39 & 5 & Average\\
ChatGPT 3.5 Turbo & 1.08 & \textbf{0.69} & 0.24 & 2.53 & 2 & Average\\
\hline
Nothing & \textbf{1.11} & \textbf{0.69} & \textbf{0.23} & \textbf{2.57} & 1 & Average\\\hline
\end{tabular}

\label{tab:gold}
\end{table*}

\subsection{Human Evaluation}

Again using the Likert scale, a human evaluator assessed the translation quality of ChatGPT and our best fine-tuned model. These translations were also evaluated in terms of adequacy and fluency. For this evaluation task, 50 sentences -- 10 from each domain -- of the Golden Collection were randomly chosen and, then, the translations of these sentences were evaluated.

Results are presented in Table \ref{tab:human}. Our best fine-tuned model had the worst performance out of all the translations, having particularly worse scores in the legislative domain. The GPT-3.5 model had similar scores to our manual reference (Table \ref{tab:humanGold}).

\begin{table*}[ht!]
\centering
\caption{Human evaluation results.}
\begin{tabular}{l|cc|cc} \hline
 & \multicolumn{2}{c}{Merged mBART-50} & \multicolumn{2}{c}{GPT-3.5 Turbo}\\
\cline{2-5}
\multirow{1}{*}{Domain} & Adequacy & Fluency & Adequacy & Fluency\\
\hline
Scientific & 4.8 & 4.7 & 5.0& 5.0\\
Legislative & 3.7 & 3.7 & 4.8& 4.5 \\
Literature & 4.7 & 3.9 & 4.6& 4.5\\
Social Media & 4.2 & 4.5 & 4.8& 4.4\\
Wiki & 4.7 & 4.5 & 4.8& 4.7\\
\hline
Average & 4.6 & 4.3 & 4.8 & 4.6\\ \hline
\end{tabular}

\label{tab:human}
\end{table*}

We can observe in Table \ref{tab:human} that our fine-tuned model fared relatively worse in comparison to ChatGPT particularly in the legislative domain. One reason why these bad scores happened could be seen in the first example of Table \ref{tab:translation_examples}, where our model oversimplified too much the information from the original sentence when translating. This ultimately resulted in the translated sentence not being fluent and being somewhat adequate. We conclude that some styles of writing, like the one present in the legislative corpus, may lead to these types of mistakes. Another example is the literary sentence, also present in the same table, but instead of oversimplifying the original information, this time the model translation ended the sentence abruptly. One explanation for this is that during training, our model learned to end sentences earlier than it should, based on factors such as the punctuation of the original sentence. However, it is interesting to note that ``forgetting'', truncating, and eliminating seemingly redundant information is not a problem exclusive to our fine-tuned model. Although it occurs less frequently, larger models such as ChatGPT, which we are currently evaluating, may also exhibit this same issue. An example of this is the Social media category, where all the onomatopoeias were eliminated, leading to a worse score in both in this model relative to our fine-tuned model. This eliminated information that was relevant to the overall context of the sentence; in this case, it indicated the sound of a phone ringing.

\begin{table*}[!ht!]
\centering
\caption{Golden Collection examples. F stands for Fluency and A for Adequacy.} 
\begin{tabular}{ l| p{0.75\textwidth} | p{0.04\textwidth} | p{0.04\textwidth}}
\hline
Category  & Sentence & F & A \\
\hline
\multirow{2}{*}{Legislative}  & \textbf{BP:} O SR.PRESIDENTE (Fernando Henrique Cardoso): – Está em votação a emenda. & - & -\\ 
& \textbf{Merged mBART-50:} O Sr. Presidente está em votação a emenda. & 1 & 3\\ 
& \textbf{ChatGPT:} O SR. PRESIDENTE (Fernando Henrique Cardoso): - Está em votação a emenda.  & 5 & 5\\
\hline
\multirow{2}{*}{Literary}  & \textbf{BP:} Leonardo ainda pôde dizer, olhando as estrelas que começavam a brilhar: - Como é belo o céu! & - & -\\ 
& \textbf{Merged mBART-50:} Leonardo ainda pôde dizer, olhando para as estrelas que começavam a brilhar: & 2 & 4\\ 
& \textbf{ChatGPT:} Leonardo ainda conseguiu dizer, olhando as estrelas que começavam a brilhar: - Como é belo o céu! & 4 & 5\\
\hline
\multirow{2}{*}{Social media}  & \textbf{BP:} TRII TRII TRII TRII Vejo meu celular vibrando em cima da escrivaninha. & - & -\\ 
& \textbf{Merged mBART-50:} TRII TRII TRII Vejo o meu telemóvel a vibrar em cima da escrivaninha. Vejo o meu telemóvel a vibrar em cima da secretária. & 5 & 5\\ 
& \textbf{ChatGPT:} Vejo o meu telemóvel a vibrar em cima da secretária.  & 4 & 4\\
\hline
\end{tabular}

\label{tab:translation_examples}
\end{table*}

\section{Conclusions and Future Work} \label{sec:conc}

Our best-performing fine-tuned model, both in the test set and the Golden Collection, was the fine-tuned Merged mBART-50. It outperformed ChatGPT 3.5 Turbo in our test set, which had sentences of the same domains our model was trained on (Subtitles and TED talks). However, the Merged mBART-50 was not capable of generalizing as well as ChatGPT 3.5 in the Golden Collection, despite achieving similar performance. These results align with the study performed by \cite{cortes-etal-2024-llms}, where ChatGPT outperformed their fine-tuned model on their Golden Collection. Although the enhancement was not significant when compared to training solely on subtitles, we conclude that incorporating a broader range of styles, sources, and data types is beneficial for improving translation outcomes overall. This is further evidenced by the results of solely relying on TED Talks for training, which resulted in inferior model performance.
\par For future work, we propose to extend the Golden Collection to more categories and sentences and to have a more comprehensive evaluation of each model. There is also room for more optimization at the training stage, namely, we would like to perform a grid search for optimizing the hyper-parameters. 

\bibliographystyle{splncs04}
\bibliography{custom}

%
%
%
%
\end{document}